# Generative AI and Process Systems Engineering: The Next Frontier

Benjamin Decardi-Nelson[1], Abdulelah S. Alshehri[2,3], Akshay Ajagekar[1], Fengqi You[1,2,*]

[1] Systems Engineering, Cornell University, Ithaca, New York, 14853, USA

[2] Robert Frederick Smith School of Chemical and Biomolecular Engineering, Cornell University, Ithaca, New York, 14853, USA

[3] Department of Chemical Engineering, College of Engineering, King Saud University, Riyadh 11421, Saudi Arabia

## Abstract

This review article explores how emerging generative artificial intelligence (GenAI) models, such as large language models (LLMs), can enhance solution methodologies within process systems engineering (PSE). These cutting-edge GenAI models, particularly foundation models (FMs), which are pre-trained on extensive, general-purpose datasets, offer versatile adaptability for a broad range of tasks, including responding to queries, image generation, and complex decision-making. Given the close relationship between advancements in PSE and developments in computing and systems technologies, exploring the synergy between GenAI and PSE is essential. We begin our discussion with a compact overview of both classic and emerging GenAI models, including FMs, and then dive into their applications within key PSE domains: synthesis and design, optimization and integration, and process monitoring and control. In each domain, we explore how GenAI models could potentially advance PSE methodologies, providing insights and prospects for each area. Furthermore, the article identifies and discusses potential challenges in fully leveraging GenAI within PSE, including multiscale modeling, data requirements, evaluation metrics and benchmarks, and trust and safety, thereby deepening the discourse on effective GenAI integration into systems analysis, design, optimization, operations, monitoring, and control. This paper provides a guide for future research focused on the applications of emerging GenAI in PSE.

Keywords: Generative AI, process systems engineering, large language models, multiscale

# List of acronyms

| Acronym | Meaning |
| --- | --- |
| AAE | Adversarial autoencoder |
| AI | Artificial intelligence |
| BERT | Bidirectional encoder representations from transformers |
| BLEU | Bilingual evaluation understudy |
| CLIP | Contrastive language-image pre-training |
| eSFILES | Extended Simplified flowsheet input-line entry system |
| FDI | Fault detection and identification |
| FID | Frechet inception distance |
| FM | Foundation model |
| GAN | Generative adversarial networks |
| GenAI | Generative AI |
| GNN | Graph neural networks |
| GPT | Generative pre-trained transformer |
| GRL | Generative reinforcement learning |
| GRNN | Generative recurrent neural networks |
| IS | Inception score |
| KID | Kernel inception distance |
| LLaMA | Large language model Meta AI |
| LLM | Large language model |
| LSTM | Long short-term memory |
| ML | Machine learning |

| | | |
|---|---|---|
| NARMAX | | Nonlinear autoregressive moving average with exogenous input |
| NLP | | Natural language processing |
| P&ID | | Process and instrumentation diagram |
| PaLM | | Pathways language model |
| PID | | Proportional-integral-derivative |
| PMC | | Process monitoring and control |
| PSE | | Process systems engineering |
| RLHF | | Reinforcement learning from human feedback |
| RNN | | Recurrent neural networks |
| SAM | | Segment anything model |
| SFILES | | Simplified flowsheet input-line entry system |
| SMILES | | Simplified molecular-input line-entry system |
| VAE | | Variational autoencoder |
| XAI | | Explainable AI |

# 1. Introduction

Process systems engineering (PSE), which encompasses the design, analysis, control, and optimization of processes and systems, has undergone a significant evolution in its scope, methods, and applications since its inception [1-4]. The primary challenges in each of these core PSE domains often include handling complexity, nonlinearity, and uncertainty in process behaviors. Notably, the advancements in PSE to handle these challenges have been intertwined with the developments in computing and systems technologies from its early days [1, 4]. The dawn of modern computing brought about new opportunities for complex problem-solving and simulation in PSE, transitioning from traditional methodologies to more sophisticated and efficient computational techniques. This evolution brought about unparalleled opportunities for complex simulations and process optimization, facilitating more efficient design, operations, and control of process systems [5]. As this relationship between PSE and computing has continued to deepen over time, the emergence of newer computational technologies, particularly artificial intelligence (AI), has begun to play an increasingly pivotal role in shaping the future of PSE.

The integration of AI has impacted many disciplines, profoundly changing their methodologies [6-8]. In particular, the integration of machine learning (ML) into PSE has marked a leap in the field's solution capabilities in handling the complexity, nonlinearity, and uncertainty in process systems [9-12]. For instance, the use of ML techniques in optimization under uncertainty resulted in about 32 % profits due to efficient planning and scheduling of process networks [13, 14]. AI-driven models have also been developed to predict reaction outcomes and optimize process parameters, thereby reducing trial-and-error approaches, and enhancing overall productivity [15]. Furthermore, this integration has improved process design and control, leading to more energy-efficient and cost-effective production methods in systems such as plant factories [16, 17]. The continuous evolution of AI capabilities promises to further elevate the PSE domain, offering solutions to previously intractable challenges and redefining the boundaries of what can be achieved in PSE.

The advent of AI models known as generative AI (GenAI) has opened new frontiers in the field of science and engineering [18, 19], with foundation models (FMs) like generative pre-trained transformer (GPT) [20], Stable Diffusion [21], contrastive language-image pre-training (CLIP) [22], and Gato [23] being recent notable additions. These GenAI models stand out for their ability to generate new data or simulations based on learned patterns and structure of data sets [24]. Particularly, the FMs, trained on vast multimodal datasets, can generate highly diverse outputs, ranging from text and images to complex simulations [25]. A notable example is the ability of GPT, particularly in its incarnation as OpenAI's ChatGPT, to generate human-like language and codes with high accuracy by finding similarities in the training data. This capability makes GenAI models versatile in finding and mimicking complex patterns often encountered in PSE data. Stable Diffusion is another GenAI model that specializes in creating high-quality images from text descriptions, while CLIP excels in understanding and integrating both textual and visual inputs, a critical advancement in multimodal learning. Gato, distinct in its versatility, demonstrates the potential of AI systems to generalize across diverse tasks and data types. As researchers continue to explore and expand the capabilities of GenAI models, their integration into various sectors promises not only to enhance current systems but also to pave the way for innovative solutions that were previously unimaginable.

The ability of GenAI models to generate unique data by learning complex patterns in datasets like the natural language capability of ChatGPT, makes them particularly useful in handling the main challenges in PSE which often include handling complexity, nonlinearity, and uncertainty in chemical systems. Figure 1 illustrates the potential of GenAI tools like ChatGPT to support molecular design and prediction, a core domain in PSE. Classic GenAI techniques such as autoencoders [26] and generative adversarial networks (GANs) [27] have already proven instrumental in addressing various PSE problems. Autoencoders have advanced molecular design by effectively generating close to 100 % of valid compounds in the Quantum Machines 9 (QM9) chemical database [28]. In a similar vein, GANs have been instrumental in creating synthetic data for rare or hard-to-obtain faulty scenarios with 99 % precision in induction motor fault simulations [29]. Furthermore, autoregressive models have been used to approximate the complex time-varying dynamics of a batch distillation column with less than 10 % deviation from the test data [30]. More recently, FMs, particularly large language models (LLMs), have been explored for their exceptional capabilities in generative tasks from predicting properties of materials to extracting knowledge from unstructured data [31]. A notable example is nach0, a multimodal natural and chemical FM leverages LLM for molecular generation, synthesis planning, and property prediction [32]. Although GenAI is increasingly being integrated into PSE, there is still a need for a comprehensive understanding and alignment of its role and impact within PSE.

In this paper, we aim to provide a systematic review of the state-of-the-art and provide fresh perspectives on the application of GenAI models within PSE. Our objective is to assess the potential advantages and challenges associated with integrating GenAI into PSE. The GenAI models considered in this paper included both the classic models like autoencoder, autoregressive, GAN, etc. as well as emerging GenAI models based on FMs. These FMs include but are not limited to GPT, Stable Diffusion, CLIP, Gato, Dall-E, etc. Considering the rapid advancements in GenAI, our study explores the implications of these developments for PSE, identifies effective GenAI techniques to enhance PSE methodologies, and examines the challenges including multiscale modeling, data requirements, evaluation metrics and benchmarks, and trust and safety, that must be overcome to fully leverage GenAI in this field. Throughout the paper, we focus on core PSE domains, which include synthesis and design, optimization and integration, and process monitoring and control.

The structure of the paper is organized as follows: Section 2 offers a compact overview and discussion of various GenAI models, including the most cutting-edge ones. Sections 3, 4, and 5 review existing studies on GenAI applications in core PSE domains: synthesis and design, optimization and integration, and process monitoring and control, respectively. In each of these sections, we also share our perspectives on the role of GenAI in PSE. Section 6 identifies and examines the major challenges that need addressing for the effective adoption of GenAI in PSE. Finally, Section 7 concludes the paper with a summary of key points and outlook.

**Example 1: Candidate crystallization solvents**

**Prompt:** Using a list of commonly used crystallization solvents for Mercaptobenzothiazole, design a novel molecule that achieves lower toxicity.

**Response:** Consider alcohol with a complex, branched structure to enhance solubility and minimize volatility, coupled with reduced chlorine content to lessen toxicity. An example of such a molecule is 2-ethylhexan-1-ol.

**Example 2: Liquid crystal reorientation**

**Prompt:** Based on the image provided, how do the molecular structures of TEA12 lead to the reorientation of 5CB liquid crystal at the aqueous interface?

**Response:** The combined hydrophilic and hydrophobic nature of TEA12 at the interface with 5CB leads to the reorientation of LC molecules, changing their alignment from parallel to perpendicular at the interface and altering the optical appearance of the LC film.

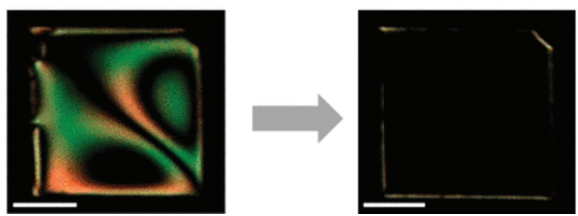

Figure 1. Examples demonstrating the zero-shot learning capabilities of ChatGPT in molecular design and prediction. In the first example, the ChatGPT provides a potential candidate molecule for Mercaptobenzothiazole crystallization solvent with lower toxicity. In the second example, ChatGPT predicts how the molecular structure of TEA12 leads to the reorientation of 5CB liquid crystal at the aqueous interface. The input image was obtained from Wang et al. [33]. While the outputs from the ChatGPT can be used as a potential starting point, experimental verification and extensive domain expertise are needed to ascertain the validity of the outputs.

## 2. Overview of Generative AI

This section introduces key GenAI methods that are applicable to PSE, highlighting their core principles, developmental progress, and associated challenges. GenAI represents a class of AI models that have the capability to create new and original data by mimicking the underlying patterns in the training datasets. Figure 1 presents an overview of the GenAI models, which are categorized into five classes: autoencoders, autoregressive models, GANs, diffusion and flow-based models, and FMs. The FMs are further categorized based on either the generation tasks (text and images) or multimodality or training framework. FMs, including notable examples such as GPT, and Stable Diffusion, are the driving force behind recent advancements in GenAI. This overview aims to provide a clear and concise summary, rather than in-depth mathematical explanations. Readers seeking more detailed mathematical insights are directed to these reviews and tutorials [24, 34-38]. This high-level overview facilitates an understanding of the relevance and adaptability of GenAI methods across diverse PSE contexts, balancing detail with accessibility.

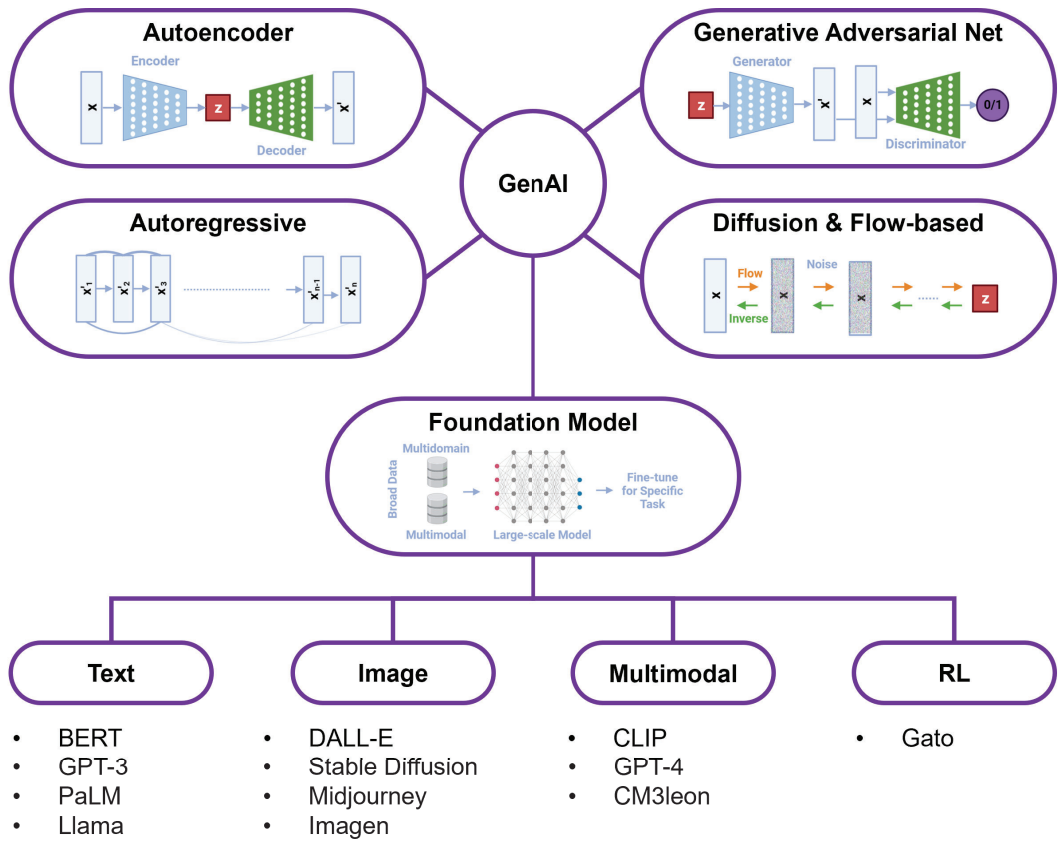

Figure 2. High-level overview of principal GenAI techniques/models namely autoencoders, GANs, diffusion and flow-based models, autoregressive models, and emerging FMs, highlighting the fundamental methods and their structures. Considering the broad base of FMs, they are further divided based on the tasks they can achieve (text or image generation), or the input data types (multimodal) or training framework (reinforcement learning).

We begin with autoencoders, dissecting their architecture and examining the advancements in variational and adversarial variants. Thereafter, we focus on GANs, scrutinizing their distinct architecture and addressing the training challenges they encounter. The discussion then shifts to diffusion and flow-based models, shedding light on their operational principles and their synergies with other model types, highlighting Stable Diffusion as a notable advancement. Thereafter, we present an in-depth examination of autoregressive models, underscoring their significance in structured data generation and recent multimodal capabilities. The exploration concludes with FMs which power many of the emerging GenAI models, focusing on the generation tasks like text and image, input data types like multimodality, and training frameworks like reinforcement learning (RL). These sections are designed to deepen the understanding of these models, tracing their evolution, and underscoring their potential role in enhancing PSE practices.

*2.1. Autoencoders*

Originating from the early 1990s [39], autoencoders consist of two interconnected neural networks: the encoder, which compresses input data into a low-dimensional latent vector, and the decoder, which reconstructs the original input from this vector [40]. Central to the functionality of autoencoders are latent variables, which are unobserved or hidden variables in the low dimensional latent vector that influence the observed data, capturing the essential features needed for accurate reconstruction [41]. The creation of latent variables, however, limits the interpretability of these models making it difficult to understand which features are more important. This characteristic can be seen as a disadvantage, particularly in PSE where interpretability can be crucial. To mitigate overfitting and address discontinuities inherent in original autoencoders, variational autoencoders (VAEs) were introduced [42]. VAEs enhance the latent space by substituting discrete points with probabilistic distributions, thereby regularizing them. The essence of VAEs lies in sampling from latent variables that likely generated the input, utilizing these samples to estimate the input probability during training. This approach provides VAEs with exceptional generalization capabilities, depicting inputs as continuous probability distributions rather than fixed points [43]. VAEs have been applied effectively in areas such as fault diagnosis within PSE, demonstrating their utility in complex system modeling and dimensionality reduction [44, 45]. Complementing VAEs, adversarial autoencoders (AAEs) leverage the GAN framework as a variational inference mechanism for latent variables [46]. AAEs thereby align the latent space distribution with a prior distribution. Furthermore, more recent advances focus on learning disentangled representations in VAEs. The objective here is to ensure each dimension of the latent vector independently encodes distinct data attributes or factors, enhancing interpretability and control in generative modeling [47, 48]. This feature positions VAEs among the more popular GenAI methods used in PSE due to their enhanced control over generative processes.

*2.2. Generative adversarial networks*

Building on the concept of adversarial training [49], GANs proposed in 2014, do not explicitly compute the probability (likelihood) of the observed data [50]. Instead, the architecture of GANs comprises two main components: the generator and the discriminator. The generator's objective is to produce data points that are indistinguishable from actual data, aiming to deceive the discriminator. Conversely, the discriminator's role is to differentiate between the generated and real data. This framework creates a dynamic of adversarial training, where the generator improves its data generation in response to the discriminator's evolving ability to identify synthetic data [40]. A distinctive feature of GANs is their construction of a latent space, simplifying input representations into a compressed form that is shared across the input domain. However, training GANs can encounter issues such as modal and model collapse. Modal collapse refers to a situation where the generator network in GAN starts to produce identical outputs leading to a lack of diversity in the generated samples. Model collapse on the other hand, refers to a situation where the discriminator is unable to distinguish between generated and real data, leading to a failure in the adversarial training process. These limitations highlight the challenges in employing GANs within PSE, where diversity and accuracy of generated data are important. To mitigate these issues, Wasserstein-GAN [51], employs Wasserstein distance to quantify the difference between distributions, offering improved stability in training. Another notable approach is f-GAN, which utilizes f-divergence as a metric for measuring distribution distance, a broader measure that encompasses KL divergence and Jenson-Shannon divergence [52]. Subsequent developments in GAN architecture, such as StyleGAN [53, 54], incorporate inductive biases which prioritize generations with specific properties and have gained widespread application in various design challenges. In PSE, GANs have been effectively used to construct the distribution of uncertainty sets for robust optimization [55]. These advanced models enhance the capacity of GANs to generate more refined and realistic data, reflecting a continuous evolution in generative modeling methodologies. The evolving capabilities of GANs make them one of the more frequently utilized GenAI methods in PSE, particularly for their ability to generate complex, high-fidelity models.

## 2.3. Diffusion and flow-based models

Diffusion [56, 57] and flow-based [58, 59] models, often grouped together due to their similarities, differ primarily in their operational mechanisms. Diffusion models adopt a stochastic approach, systematically adding noise to the data and subsequently learning to reverse this process for data reconstruction [24]. Delving deeper, diffusion models draw inspiration from non-equilibrium thermodynamics. The process consists of forward and backward diffusion processes, where the forward phase gradually introduces noise, changing the data into a Gaussian distribution. Conversely, the backward phase diminishes this noise, reconstructing the original data from its noisy counterpart. Despite their ability to capture detailed data features independently from the data distribution, diffusion models face challenges due to the slow addition of noise, leading to lengthy training times [36]. This challenge may limit their immediate applicability in PSE applications where faster model training and iteration are typically required.

Flow-based models, alternatively known as normalizing flows, leverage invertible deterministic transformations between data and latent spaces. Their standout feature is the direct computation of data likelihood, which is particularly useful for explicit density estimation. However, these models are constrained by their dependency on complex hyperparameter tuning and the necessity to change input data into a continuous form for optimal performance [24]. This limitation can pose practical challenges in PSE, where data often exhibit a wide variety of distributions and discontinuities. In recent developments, efforts to integrate diffusion with VAEs [21], GANs [60], flow-based [61], and autoregressive [62] models have shown promising outcomes, surpassing the capabilities of individual generative models. The emergent hybrid diffusion approaches combine the strengths of different model types while addressing sampling inefficiencies of diffusion models. This synergy not only enhances model representativeness but also facilitates the generation of complex distribution features beyond the reach of individual models [24]. Hybrid models are desirable in PSE due to their enhanced capability to handle complex, multi-modal data distributions.

### 2.4. Autoregressive models

First introduced in the 1920s [63], autoregressive models have retained their effectiveness. They adeptly distill complex variables into conditional probabilities, a method that remains important for achieving state-of-the-art performance in contemporary language understanding tasks. Their methodical approach lies in sequentially predicting each variable component based on prior elements, effectively maximizing data likelihood and diminishing negative log-likelihood [64]. A hallmark of autoregressive models is their exceptional capability for density estimation. This advantage, however, is tempered by slower processing in high-dimensional data and the requirement for fixed data sequencing, which can influence performance outcomes [24]. This limitation is particularly relevant in PSE, where the complexity and scale of systems can hinder the utility of models requiring sequential data processing. Architectural progress in autoregressive models is geared towards expanding their receptive fields and memory. Initial strategies involved techniques like masking multilayer perceptron autoencoders, exemplified by the neural autoregressive density estimator [65]. Similarly, recurrent neural networks (RNNs), though initially developed, have been adapted to autoregressive models to better manage extensive data ranges [66]. The most noteworthy advancement in autoregressive models is the integration of self-attention mechanisms, notably in Transformer models [67]. These mechanisms enable selective focus on sequence segments at each generation step, enhancing parallel processing, stable training, and the learning of distant dependencies, especially evident in recent advancements in natural language processing (NLP). Overall, these innovations have broadened the scope and applicability of autoregressive models. In PSE, autoregressive models are particularly valued for their precision in time-series forecasting, making them a common choice for modeling dynamic systems [68]. Such models also play an important role in generating structured data, including images [69], and graphs [70], and are increasingly relevant in addressing several problems within PSE.

*2.5. Foundation models*

FMs in GenAI represent a significant shift in the field of AI. Historically, AI research focused on specialized models tailored for specific tasks. However, with the advent of models like GPT and its successors, the field witnessed a shift. FMs are large-scale models, trained on vast datasets, and can generate highly diverse outputs, ranging from text and images to complex simulations [25]. The ability of FMs to learn from a broad spectrum of data enables them to adapt and respond to a wide range of tasks and queries, often with a high degree of accuracy and creativity. This ability makes them particularly adept in few-shot and zero-shot learning situations allowing FMs to generalize on specific tasks with only limited training samples. FMs often make use of several classic GenAI techniques coupled with training on large datasets for improved performance. The term "foundation model" highlights their role as a base upon which various AI applications can be built, reflecting their versatility and potential in the field of GenAI. In this section, we provide an overview of the FMs for text and image generation, multimodal FMs, and FMs based on RL. We introduce a subsection for RL to distinguish the training framework from the standard training framework typically used in GenAI.

*2.5.1. Text generation*

The recent advent of LLMs like the GPT series marks a major leap in natural language understanding, evolving through four stages: Statistical Language Models [71], Neural Language

Models [66, 72], Pre-trained Language Models [20, 73], and LLMs [20, 73]. These models, especially GPT-3 with its 175 billion parameters, have showcased remarkable adaptability in zero-shot and few-shot learning scenarios across various NLP tasks [73]. The integration of RL from human feedback (RLHF) in models like InstructGPT (ChatGPT) has refined their alignment with human-like conversation and understanding [74]. This evolution culminated in GPT-4, the current state-of-the-art model, introducing multimodal capabilities to process both text and images, significantly enhancing its versatility and applicability in complex scenarios [20]. Another model is BERT (Bidirectional Encoder Representations from Transformers) [75], developed by Google. BERT, utilizing the Transformer, stands out for its approach to contextual word representations. Unlike previous models that processed text in a unidirectional manner, BERT analyzes words in relation to all the other words in a sentence, allowing for a more nuanced understanding of language. This has led to substantial improvements in tasks like sentiment analysis, question-answering, and language translation. Although BERT's contextual understanding is advantageous, its reliance on large context windows can lead to inefficiencies in processing shorter texts, a challenge in certain PSE applications like molecular design.

Other notable GenAI FMs include Google's PaLM (Pathways Language Model) [76] and OpenAI's LLaMA [77]. PaLM, with its scalable architecture, represents a step forward in efficiency and generalization. The architecture of PaLM is built on a large-scale Transformer model, enabling it to efficiently process a vast amount of information and learn from diverse data. PaLM excels in multitasking and can handle a variety of language tasks without the need for task-specific tuning, demonstrating remarkable performance in areas like code generation and language translation. LLaMA, similarly, is an example of the ongoing evolution in language models. The architecture of LLaMA also focuses on leveraging a large Transformer-based model but with a more efficient use of parameters. LLaMA's parameter efficiency is especially beneficial in PSE, where computational resources can be a limiting factor. This efficiency allows LLaMA to achieve high performance with fewer resources, making it a notable development in optimizing the balance between model size and computational efficiency.

### 2.5.2. Image generation models

GenAI FMs for image generation represent a shift in the field of AI, blending creativity and technology in unprecedented ways. These models are typically either diffusion-based or GAN-based. Among the most notable is OpenAI's DALL-E [78], introduced in 2021, which pioneered the concept of generating detailed images from textual descriptions. This model leverages a modified GPT-3 architecture to understand and visualize a wide array of concepts, from surreal art to photorealistic images, demonstrating a significant leap in AI's ability to understand and create visual content. However, the challenge with DALL-E in practical applications such as PSE is managing the balance between creative freedom and the accuracy required for scientific images. Stable Diffusion [21], launched by Stability AI in 2022, became another key player. Using VAE, U-Net, and optional text encoder, Stable Diffusion offers a unique approach to generating high-quality images, allowing for more control over the creative process. This capability of Stable Diffusion to fine-tune image attributes makes it particularly useful in scenarios where precision and customization are required, such as in molecular design.

Another remarkable advancement in this domain is Midjourney, which emerged in early 2022. Midjourney stands out for its ability to create highly artistic and stylistically varied images, catering to a broad spectrum of aesthetic preferences. The versatility and ease of use of Midjourney have made it a popular tool among artists and designers. Additionally, Google's Imagen [79], revealed in 2022, represents another significant stride. The architecture of Imagen is built on a Transformer-based model, enhanced with techniques for achieving exceptional photorealism and detailed text-to-image generation capabilities. This setup allows Imagen to interpret and visualize textual descriptions with impressive accuracy and lifelike detail, setting new standards for image quality in AI-generated art.

### 2.5.3. Multimodal models

Multimodal models signify a major leap in AI, seamlessly blending diverse data types such as text, images, graphs, and audio into comprehensive models [80]. Surpassing traditional single data models, multimodal models adeptly process and produce multimodal content. A notable example is GPT-4 which powers the latest iteration of ChatGPT. Another is Google's CM3Leon [81]. CM3Leon is a retrieval-augmented, multimodal language model based on the CM3 architecture [82], uniquely trained using a method adapted from text-only models, capable of generating both text and images. CM3Leon achieves state-of-the-art performance in text-to-image generation with significantly less training computational resources and demonstrates exceptional controllability in various tasks like image editing and generation. Multimodal models evolve through two primary categories, often involving: converters, altering multimodal data for compatibility, and perceivers, enhancing the perception of the model across various modalities [83]. Despite their advancement, challenges in bridging semantic gaps can lead to problematic outcomes, like erroneous content generation, and inefficient modality alignment, escalating computational costs for marginal performance gains [84]. In PSE, these challenges could affect the accuracy and efficiency of simulations and predictions, making careful alignment and integration techniques necessary.

Converter multimodal models skillfully merge various data types, tackling the challenge of modality-specific semantic differences. Their development is characterized by three distinct approaches. Direct Mapping, exemplified by LLaVA [85], FROMAGe [86], and VisionLLM [87], primarily utilizes fundamental image encoding techniques, such as CLIP [22] vision encoder, and projection methods. This approach simplifies integration but often results in alignment issues and hallucinatory errors. Textual conversion, represented by models like PICa [88] and REVIVE [89], converts visual inputs into text, thus translating visual data into a format more comprehensible for LLMs. While effective in bridging the semantic gap, this method can oversimplify visual information, a significant drawback in detailed PSE tasks requiring high fidelity visual data representation. Adapter-Based adjustment, seen in Llama-Adapter [90, 91] and Prophet [92], introduces encoded image features as adjustment layers, enhancing the LLMs' grasp of specific tasks. Each strategy offers unique benefits: Direct Mapping for its straightforwardness, Textual Conversion for enhancing comprehension, and Adapter-Based Adjustment for its depth of understanding [83]. However, these methods also face challenges, including semantic misalignments and an excessive dependence on LLMs' learning capabilities, paving the way for alternative multimodal variants.

Perceiver multimodal models are engineered to narrow the semantic divide between text and other data forms, such as images and audio. These models adopt three primary types of perceivers: VAE Perceivers, Q-former Perceivers, and Customization Perceivers. VAE Perceivers, like BEiT [93] and LQAE [94], employ VAEs for creating visual tokens, aligning image and text modalities. Q-former Perceivers, including BLIP-2 [95], Stable Diffusion [21], and MiniGPT-4 [96], utilize query embeddings and image features in a Transformer-based structure, enhancing image-text matching and generation. Customization Perceivers, such as Flamingo [97] and VisionLLM [87], offer specialized modules for adapting multiple visual modalities into language model-friendly formats. While these approaches enable more cohesive multimodal understanding, they face challenges in maintaining alignment accuracy, particularly when managing complex and diverse data types. This necessitates ongoing innovation in perceiver designs and methodologies for improved multimodal integration.

Current research in multimodal models concentrates on evolving more complex perceiver models, especially those akin to the Q-former architecture [83]. Despite their nascent stage, multimodal models show promise, yet demand further architectural development. Present models, primarily designed for specific modalities like images, face challenges in concurrently handling diverse modalities. This limitation is particularly important in the PSE field, where modalities like graphs are important. Thus, a unified model adept at managing multiple modalities simultaneously remains a key area for future exploration and innovation.

### 2.5.4. Reinforcement learning models

RL, an approach conceived in the 1980s, is a paradigm of ML where an agent learns to make decisions through trial-and-error interactions within an environment to achieve specific goals [98]. In this process, the agent receives feedback in the form of rewards or penalties, learning to adopt strategies that maximize cumulative rewards. Unlike traditional ML methods that primarily focus on pattern recognition, RL is centered around learning optimal decision-making policies under uncertainty. The connection between RL and GenAI lies in their shared capability to handle dynamic, complex problems. GenAI models, particularly in tasks that require creativity and generative processes, explore vast possibilities to generate high-quality outputs. This aspect of exploration and optimization in GenAI resonates with the core principle of RL, where agents explore and learn the best actions to maximize rewards. This common ground makes RL techniques invaluable for enhancing LLM models, especially in areas requiring adaptive decision-making.

DeepMind's MuZero [99], introduced in 2020, is a landmark in the RL field. MuZero uniquely combines model-based and model-free RL approaches, enabling the agent to master complex games like chess, shogi, and Go without pre-existing knowledge of their rules. In 2022, DeepMind further bridged the gap between RL and GenAI with the development of Gato [23]. This multi-modal, multi-task model can perform over 600 tasks across diverse domains including text generation. Gato exemplifies the integration of RL in GenAI, allowing the model to not only generate content but also to interact effectively in various environments, from gaming scenarios to robotic control tasks. The adaptability of Gato highlights its potential in PSE for applications like process control and optimization, where versatile and adaptive decision-making is critical. Both MuZero and Gato mark the synergistic potential of combining RL with GenAI, illustrating how RL's strengths in decision-making can be leveraged to enhance the generative and interactive capabilities of GenAI models. This fusion marks a significant step towards more adaptive, intelligent, and versatile AI systems, capable of creative generation and intelligent environmental interaction, propelling the field of AI into new possibilities. However, a major limitation is the inability of these models to guarantee constraint satisfaction in safety-critical systems.

## 3. Generative AI for Multiscale Systems Design

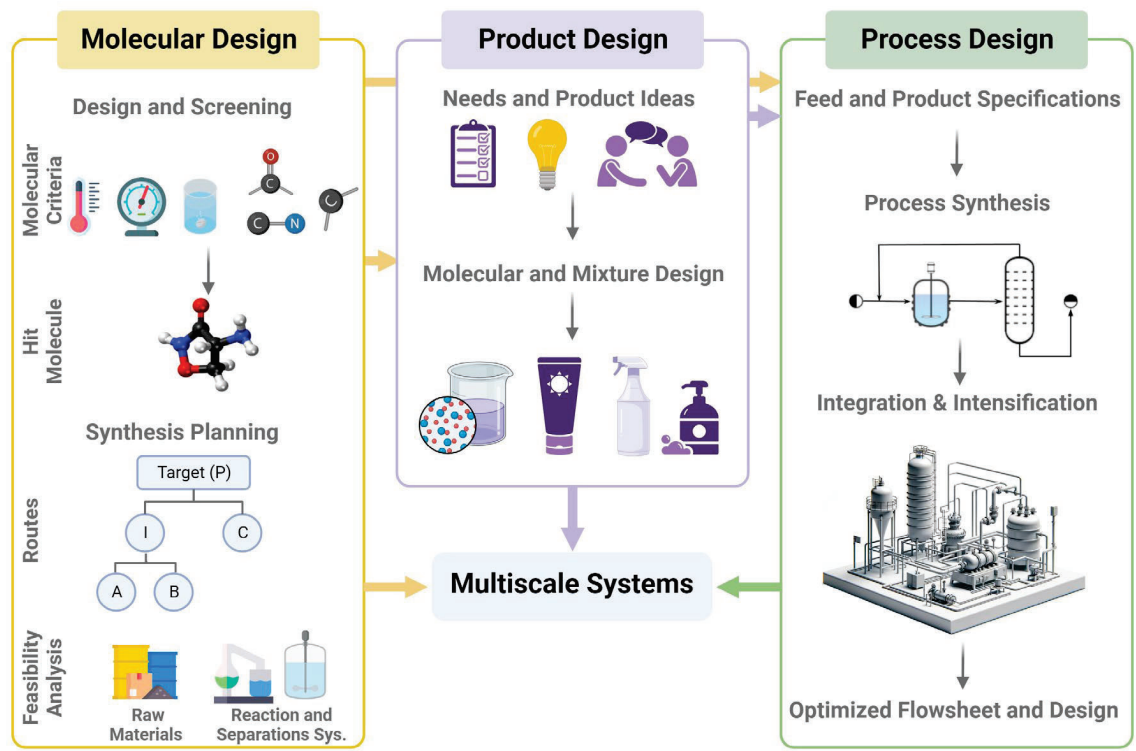

Figure 3. Detailed illustration of the role of GenAI in facilitating molecular, product, and process design, with emphasis on the interconnections and interactions as indicated by arrows.

In PSE, the triad of design– encompassing molecular, product, and process design – stands as a cornerstone for creating novel chemical structures, products, and processes [100]. These distinct yet interconnected facets, as illustrated in Figure 2, represent a range of design challenges and their interactions at varying scales. From a systems view, this interconnectedness is pivotal, underlining a holistic approach to PSE design that synergizes individual components for optimized overall performance. At the most granular level, molecular design aims to computationally generate molecules or modify existing ones to endow them with specific properties and functionalities, addressing critical needs in various applications [101, 102]. In contrast, product design in PSE encapsulates end products lifecycle from conceptualization to manufacturing, translating market needs into functional, market-ready products [103]. Process design, on the other hand, involves the systematic development and optimization of process flowsheets, necessary for converting feedstocks into desired chemicals and products [104, 105]. It should be emphasized that these design problems are deeply interwoven, particularly with molecular design being a cornerstone within the broader matrix of product and process design [106, 107]. This interdependency accentuates the multiscale complexity of these challenges, reflecting a systems-oriented paradigm that requires a comprehensive understanding of how individual elements interact within the larger system.

The PSE design domain has been notably driven in the past by major strides in computing hardware and computational methods [2]. Incorporating a systems view, these advancements have enabled a more integrated approach in PSE, where computational capabilities are harnessed to optimize the entire range of design challenges. This development is evident in the array of methodological innovations, particularly in mathematical optimization [108-110] and stochastic metaheuristics [111, 112]. These areas have seen active contributions from PSE research, particularly in developing computational algorithms [113-115]. Recent and central to this evolution is the emerging role of GenAI methods, which have recently changed inverse design in general by synthesizing novel designs through implicit learning from an extensive array of existing designs, subject to both explicit and implicit constraints [35]. In the chemical space, GenAI has demonstrated exceptional prowess, surpassing traditional methods in generating high-performance drugs [116, 117] and materials [118, 119], reaction synthesis routes [120, 121], stable 3D molecular structures [122, 123], among others. Such advancements highlight the potential of GenAI in offering system-wide solutions, transcending the limitations of isolated approaches. As we look ahead, the collaborative integration of GenAI into PSE design not only can align with the interplay of molecular, product, and process design, but also heralds a new era of innovation and efficiency in addressing the inherent complexities of these intertwined designs.

In this subsection, we discuss the adoption of GenAI as an emerging tool across the PSE design stages. We explore the early applications of GenAI, showcasing its potential to transform traditional methods and contribute to the development of innovative design solutions in PSE. It is important to emphasize that although GenAI has versatile applications within various facets of PSE design, our review is specifically tailored to its utilization as a design or synthesis method. This focus aligns with a systems-oriented approach, where GenAI's role in PSE is viewed not in isolation but as part of a larger, interconnected framework of design methodologies. We also draw parallels with other disciplines, where GenAI's application to structurally analogous problems have yielded promising results. These analogies serve to illuminate the applicability of GenAI in PSE, foreshadowing its future trajectory and potential to revolutionize the field.

*3.1. Molecular Design*

Molecular design serves as a pivotal element of chemical systems, shaping the macroscale behaviors of products and processes through the manipulation of structures and properties at the microscale. This facet of PSE design intertwines four components: molecular representation [124-126], property estimation [127-129], design methods [130], and synthesis planning [120]. The role of design methods, including GenAI, is to explore the vast expanse of chemical space intelligently and efficiently. Yet, the efficacy of such methods, particularly GenAI, hinges on a complex balance of these components. Consequently, the incorporation of GenAI in molecular design not only necessitates domain knowledge but also a nuanced understanding of the representational needs specific to the problem at hand [131]. Integral to molecular design, synthesis planning employs predictive or GenAI algorithms to identify feasible reaction sequences, thereby enhancing the efficiency and innovation in molecular design processes. This inclusion of synthesis planning within the design framework marks a critical step in bridging the gap between computational molecular design and experimental efforts [120]. Therefore, addressing PSE molecular design problems with GenAI is fundamental to harnessing its full potential, particularly in addressing the complex, multiscale nature of PSE.

Generative models have demonstrated exceptional ability as design methods to generate molecular candidates and synthesis routes [131]. Example GenAI improvements on the molecular design front include the generation of superior metal-organic frameworks with improved natural gas separation properties [132], novel drugs with tailored antibiotic activity [133], and batteries with significantly reduced lithium requirements [134]. Such models have traversed a range of generative strategies and representations, evolving from simpler character-based forms like simplified molecular-input line-entry system (SMILES) [135] to sophisticated graph-based approaches. Within PSE, despite the extensive use of ML models in molecular design as property models [136-138], GenAI applications that handle the generation end-to-end remain limited. Initial forays utilized natural language-inspired autoregressive models for generating organic structure directing agents with character-based representations [139, 140]. The transition to graph-based representations marked a significant advancement, establishing state-of-the-art results across a spectrum of molecular tasks [141]. A pioneering application of graph-based generation is the high-octane fuel design study [142], comparing the efficacy of three leading generative models: Junction-Tree VAE [143], Molecular Hypergraph Grammar VAE [144], and Molecular GAN [28]. The study highlighted the efficacy of VAE in generating high-quality molecular candidates. However, subsequent experimental validations revealed a critical need for larger physical and chemical property data to generate candidates with higher confidence in property values [142]. Further exploration in the generative design domain has led to autoregressive graph generation techniques being adapted for solvent design, specifically targeting the separation of cyclohexane and benzene [145]. Yet, a common challenge facing these methods is their heavy reliance on computational data rather than experimental data, potentially leading to data and knowledge gaps in the generated molecules [142].

In synthesis planning, due to its structured and multi-output nature, GenAI has predominantly been applied using autoregressive [121, 146] and generative RL (GRL) [147, 148] models. It is important to note that deep learning approaches have significantly improved chemical reaction classification accuracy from 50% (achieved with traditional hand-crafted methods) to 98.2%. However, this comparison is limited to a relatively simple task [149]. GenAI's true potential lies in its ability to address more complex challenges, such as the generation of reactants, products, and reaction networks [150]. Still, automating feasibility assessments remains a significant challenge within this domain, requiring accurate estimation of synthesis costs that consider various factors like steps, yields, and reactant prices [149].

Drawing inspiration from state-of-the-art approaches in the parallel fields of drug and protein design, recent trends highlight the application of advanced flow-based and diffusion models, augmented by roto-translation equivariant attention-based graph neural networks (GNNs) [151-153]. These models have demonstrated remarkable capabilities in several areas. Firstly, the models have enhanced our understanding of complex drug or ligand-protein binding interactions within the generation task [154]. Moreover, the models have established the state-of-the-art in generating 3D low-energy molecular conformers [122], notably outperforming traditional rule-based software by significant margins and avoiding the high computational cost entailed by density functional theory methods [123, 155]. Furthermore, in scenarios where conformational specifics are less pivotal, as often in PSE molecular contexts, diffusion models can offer unified 2D molecular generation [156]. Despite their enhanced performance, these models pose computational challenges including training time and sample efficiency, requiring a careful balancing act in their application, particularly when compared with VAEs and GANs [157]. Nonetheless, integrating such techniques within PSE holds the potential to improve the quality of generated candidate molecules leading to computational discoveries of novel chemicals, solvents, fuels, among others.

Recent advancements have seen LLMs being employed in molecular design, demonstrating promise despite their current limitations in regression capabilities [158]. These models are increasingly recognized for their potential in a variety of applications within molecular design. One notable application involves the improvement of textual molecular descriptions by LLMs, establishing state-of-the-art performance across several molecular properties [158]. Furthermore, LLMs have been explored for their proficiency in predictive modeling of molecular properties [159, 160], generative tasks [31], and multimodal interpretations [161]. A prime example of LLMs in molecular design tasks is nach0, which leverages an LLM for an array of chemical tasks, demonstrating state-of-the-art performance across single-domain and cross-domain tasks, including molecular generation, synthesis planning, and property prediction [32]. Such applications suggest that LLMs, with their enhanced textual analysis and feature enrichment capabilities, could surpass current generative models and paradigms in the field of molecular design.

### 3.2. Product Design

Product design in PSE is a multifaceted and multilayered problem, extending from initial ideation and understanding customer needs to the processes of manufacturing. This design sphere in PSE covers a diverse array of products, ranging from individual species to mixtures, as well as formulated products and functional devices [103]. A fundamental hurdle in this design problem is its multiscale dimension, necessitating precise representation of products and the nuanced interactions across different stages. Such complexity is amplified when linking microscale properties to macroscale performance outcomes in model-based product design [162]. Additionally, PSE product design demands a balance of various stringent criteria, encompassing sustainability, cost-effectiveness, and compliance with safety standards [103, 107].

Although GenAI offers substantial potential to supplant the current heuristic and Edisonian approaches, its integration into product design remains limited. Yet, valuable insights gleaned from successful applications of GenAI in fields such as materials science, and transportation sectors, provide a blueprint for innovating chemical product design. In materials science, GenAI has facilitated the generation of advanced microstructures and metamaterials topologies, utilizing GANs [163, 164], VAEs [165, 166], and flow models [167]. In parallel, the automotive and aerospace sectors have pioneered the synthesis of 2D and 3D shapes for aerodynamic efficiency with GRL and GANs in single [168, 169] and multi-component systems [170, 171]. These advances can be translated to PSE, particularly in synthesizing product properties and geometries, and optimizing shapes within complex chemical or physical simulations. Further, emerging multimodal autoregressive models in material design have shown remarkable capability in advancing natural language descriptions into materials 3D structures [172]. Within PSE product design, these models can effectively bridge the gap between customer requirements and tangible product designs. Furthermore, such models open up avenues for tailoring chemical products to specific needs, regulatory requirements, and customized functionalities [173]. The success of GenAI in these closely related fields could be replicated in PSE to improve the solution methodologies.

In product design, no single method reigns supreme due to the diverse nature of design tasks. Yet, recent advancements in multimodal autoregressive models, particularly in material design, demonstrate an impressive ability to convert natural language descriptions into 3D material structures [172]. These models are increasingly recognized for their effectiveness in generative design methods, effectively bridging the divide between customer specifications and concrete product designs. This development signifies a pivotal shift in how product design aligns technological innovation with user-centric needs.

### 3.3. Process Design

Process design in PSE involves three interrelated stages: flowsheet synthesis for efficient flowsheet creation; integration for resource optimization; and intensification to improve performance and sustainability [174]. This section focuses on flowsheet synthesis, while integration and intensification are addressed in subsequent discussions. In flowsheet synthesis, molecular and product design are integral, influencing all aspects from raw material selection to evaluating process feasibility and economic viability [100]. Challenges here encompass developing and standardizing process flowsheets for optimal efficiency, addressing synthesis intricacies, and overcoming limitations in current representation methods and software tools [175]. Recent GenAI applications to process synthesis have shown promising results, particularly in two directions: process flowsheet representation and flowsheet completion and generation.

Representation methods serve as a foundational basis for generative tasks. This is particularly important in sequential processes such as process synthesis, where each decision sets the stage for the next, creating a cascading effect on the overall outcome [175]. Early process flowsheet representation methods utilized fixed-size matrices and tensors in conjunction with RL, serving a non-generative optimization strategy [176, 177]. These representations typically encompass five matrices: molar flow and fractions, unit types, unit connectivity depicted as an adjacency matrix, and task termination criteria [177]. Yet, such a representation decision primarily stems from the straightforward implementation of the representations into optimization algorithms [175]. Alternatively, drawing from textual molecular representations like SMILES, the simplified flowsheet input-line entry system (SFILES) was developed, introducing a text-based notation for chemical process flowsheets [178]. Subsequent enhancements in SFILES 2.0 have expanded its capabilities to encompass the full topological information [179]. Further, the extended version, eSFILES, integrates a hierarchical text and graph representation across four levels: flow diagrams, connectivity, material and energy balances, and relational ontologies [180]. However, the integration of multiple data levels and modalities can complicate the generative tasks. Consequently, recent flowsheet generation works have shifted towards graph-based models [181, 182], closely mirroring the inherent structure of chemical flowsheets.

In terms of process flowsheet generation, several generative models have shown early promise. Autoregressive models, specifically, have been utilized for the completion [183, 184] and correction [185] of process flowsheets, demonstrating varying degrees of success. A notable limitation of these generative models is their dependency on extensive datasets and the lack of thorough simulation and evaluation of the generated processes [175]. To address these challenges, RL methods have been extensively employed to both generate flowsheets and assess the designs, utilizing either approximation methods or rigorous simulation environments such as COCO [186], Aspen Plus [176, 187], UniSim [188], and DWSIM [182]. This evaluation process progressively refines the quality of generated designs, guiding the RL agent toward improved flowsheets. Highlighting the potential of RL in process design and simulation, one study demonstrates that GenAI-based approaches outperform exact optimization methods in flowsheet design by 16% [181]. Despite the promise of RL methods in enhancing process flowsheet generation, a number of implementation challenges exist. These include complex setup, data inefficiency, high computational demands, and potential issues with convergence during training [189]. Although some RL applications have explored pretraining techniques to enhance data efficiency [182, 190], they still face significant hurdles, such as slow learning times and a limited ability to utilize existing data, impeding their efficacy [175]. Therefore, merging generative models with RL approaches could potentially enhance the efficiency and knowledge transfer in process flowsheet design. To date, no existing literature has combined both GenAI and RL to advance the current state of process design. Furthermore, exploring the potential of Gato, a generalist RL agent, to advance flowsheet generation is a promising direction. Another direction is the use of LLMs such as ChatGPT to understand eSFILES efficient flowsheet generation. These directions are currently underexplored.

## 4. Generative AI for Optimization

Optimization is at the heart of PSE which intersects mathematics, engineering, and computer science to enable solutions to complex problems at various scales [191]. Advancements in computing and software engineering have had a significant impact on the field of optimization transitioning from an academic research interest into an effective instrument with industrial utility. For instance, in the evolving field of industrial operations, advancing from expert intuition to modeling and optimizing complex systems–driven by improvements in optimization algorithms– is key to achieving resource efficiency, cost-effectiveness, and environmental sustainability [109]. Optimization in PSE encompasses a wide array of challenges, including the need to guarantee robust performance under varying conditions, particularly for real-world systems which often exhibit uncertainties, nonlinearities, and discrete decision variables, making the optimization task complex and multifaceted [192]. Efforts in PSE have expanded beyond deterministic optimization, recognizing the importance of accounting for uncertainties inherent to problem parameters. The growing complexity and scale of process engineering models incorporating uncertainty has driven the development of large-scale optimization algorithms by addressing factors like problem size, explosive growth of combinatorics with discrete variables, and the effects of underlying problem structure [193]. The quest for more effective optimization strategies in PSE has paved the way for the integration of GenAI and its associated versatility and adaptability capabilities into the optimization domain.

Although optimization algorithms for addressing a variety of problem formulations have been put forth through mathematical underpinnings of theoretical foundations and rigorous formalism [194, 195], mathematical intuition has been recently shown to benefit from GenAI-based structured guidance [196]. By harnessing the power of GenAI, optimization in PSE can also endeavour to push the boundaries of efficiency, reliability, resilience, and adaptability, while addressing large-scale problems associated with complex process systems. In this section, we identify and explore specific areas where GenAI has seen some disruption and exhibits revolutionary potential to affect optimization in PSE. An overview of these areas along with their components that can be influenced with GenAI are shown in Figure 3. Integration of GenAI can offer data-driven variants for *optimization under uncertainty* while enhancing the robustness of optimal solutions. Generative deep learning models also emerge as a promising tool for *learning* *to optimize* by improving exploration of solution space expressed by combinatorial optimization problems commonly arising in PSE. GenAI techniques further offer avenues for leveraging deep learning models in *surrogate-based optimization* for complex process systems by aligning data-driven surrogates with decision-making in PSE. Lastly, GenAI not only enables efficient optimization of computationally intensive PSE problems but also presents opportunities for *metaheuristic optimization* to train deep generative models.

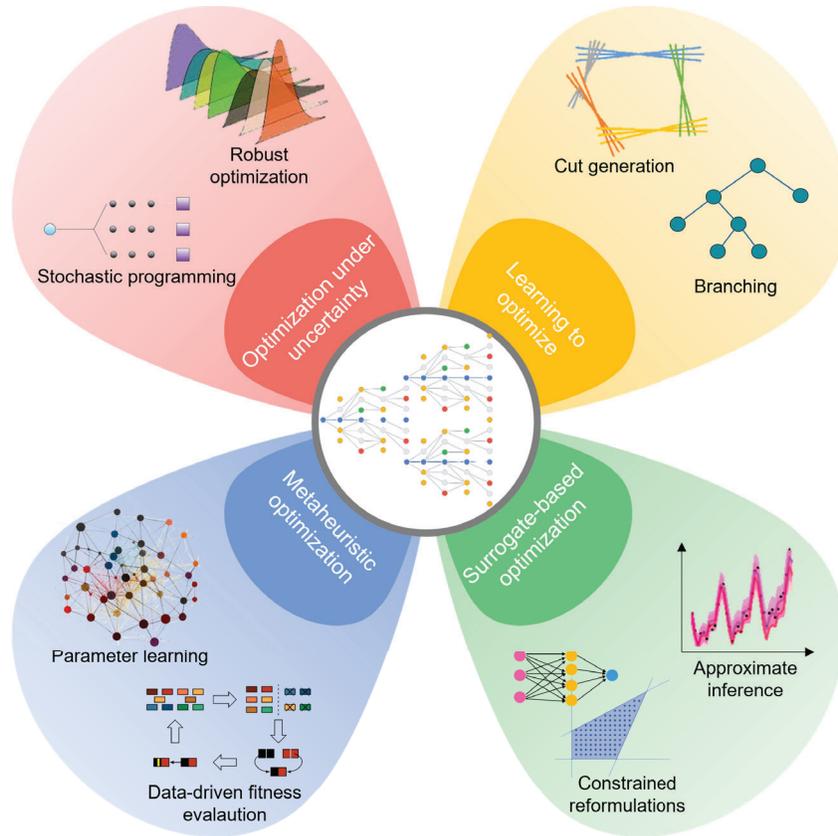

Figure 4. Components of optimization in PSE that exhibit the potential of utilizing GenAI for improvised performance capabilities. These comprise tackling optimization problems under uncertainty, leveraging learning for the optimization process, using generative deep learning models as surrogates, and incorporating learning in metaheuristic optimization algorithms.

### 4.1. *Optimization under uncertainty*

As the variability of parameters can largely influence the solution to an optimization problem, deterministic problem formulations are generally not suitable for practical applications in PSE [197]. Owing to the need for robust and reliable design and operation of process systems, optimization under uncertainty has been an active area of research [198-200]. Mathematical programming techniques like stochastic programming [201], robust optimization [202], and chance-constrained programming [203] are commonly adopted to address optimization problems that incorporate uncertainties into their mathematical models. Stochastic programming leverages a scenario-based approach wherein decision making is performed under potential realizations of uncertain parameters following a known probability distribution [204], while chance-constrained programming deals with constraint satisfaction of an optimization problem in a probabilistic manner [205]. On the other hand, robust optimization techniques model uncertain parameters using an uncertainty set to hedge against the worst-case realization and does not require accurate knowledge of underlying probability distributions [206].

The rise in the adoption of ML in PSE has introduced fertile avenues for data-driven variants of these traditional approaches for optimization under uncertainty [207]. Data-driven optimization strategies have witnessed an increase in use of deep learning models as tools for hedging against uncertainty and can benefit even further from sophisticated GenAI methods. Generating scenarios with GenAI methods by learning the data distributions can be utilized by scenario-based optimization in stochastic programming to benefit from synthetic uncertainty data for increased precision in estimating expectation values [207]. GenAI techniques like GANs have been extensively used for future uncertainty realization predictions [208-210]. Even variations of GANs like Bayesian GANs [211] and other GenAI techniques like VAE [212, 213] have been exploited to generate scenarios that not only capture the statistical characteristics of historical uncertainty realizations but also create new generation patterns [214]. Applications of deep generative models used as a sampling tool for scenario-based optimization can contribute to high-quality solutions [215] which can be leveraged in PSE to develop integrated stochastic programming and GenAI approaches [216] for optimization under uncertainty. However, GenAI methods may not always generate reliable scenarios that represent real-world uncertainty primarily due to limiting factors of significant historical data requirements for training sophisticated architectures like GANs [217]. The limited interpretability in complex deep learning models may require plausibility evaluation of GenAI-generated scenarios with domain expertise. Owing to the capabilities of capturing and generating probabilistic information, LLMs can also be potentially leveraged for explicit uncertainty modeling. With unknown probability distributions of uncertain parameters, GenAI methods like GANs can assist in generating ambiguous probability distributions to develop distributionally robust chance-constrained programming frameworks for systems optimization [218]. Provisions to tackle multiple chance constraints using distributional information extracted from GANs can also be accommodated in such frameworks [55]. Models like ControlNet and Stable Diffusion provide control over image generation through conditioning [21, 219]. Such diffusion-based GenAI models can be applied to manage information flow while quantifying uncertainties from data recorded in various PSE operations and aid in preventing fallacies associated with training deep learning models. These fallacies include overfitting, where a model learns the training data too well, including its noise and outliers, leading to poor generalization to new, unseen data. Another fallacy is underfitting, where the model fails to capture the underlying trends in the data, resulting in inadequate performance even on the training data. Additionally, these models can help avoid the fallacy of data bias, where the model develops biases based on unrepresentative or prejudiced training data, leading to skewed or unfair predictions. Despite the potential of using sophisticated methods to overcome fallacies associated with integrating deep learning into optimization under uncertainty for PSE, the computational costs of training and running GenAI models can limit the scalability and real-time applicability of stochastic optimization for practical use cases [220].

Apart from learning probability distributions encapsulating uncertainty data, GenAI is also capable of performing chance constraint learning [221] to incorporate robustness into optimal solutions for process systems optimization. In addition to stochastic and chance-constrained programming, data-driven robust optimization with uncertainty sets generated by deep learning models has been shown to effectively hedge against uncertainties [222, 223]. Although the integration of deep learning models and robust optimization in PSE has been limited to unsupervised clustering [224, 225], GenAI methods can supplement the learning components in data-driven robust optimization [13] to extract hidden structures and anomalies from historical uncertainty realizations. Additionally, multimodal learning can facilitate integrating information from diverse sources and provide a comprehensive understanding of the uncertainties arising in PSE [226]. Integrating LLMs and multimodal learning with models like GPT-4 [227] can also contribute to learning representations that capture the inherent structures and patterns in uncertainty data. This can further enhance robustness to noisy or incomplete data by filtering out noise and focusing on relevant information through attention mechanism exploited by models like GPTs at various scales [228]. For optimization under uncertainty, not only conventional GenAI methods but also advanced technologies like LLMs, multimodal LLMs, and diffusion-based models offer promising avenues to model, represent, and adapt to uncertainty for robust decision-making in PSE.

### 4.2. *Learning to optimize*

Mathematical programming algorithms to solve problems in process systems optimization have been emerging and evolving subject to interplay between theoretical underpinnings, sophistication of computation, and real-world requirements [200, 229]. Solution techniques for varying formulations ranging from linear programming to mixed-integer nonlinear programming rely on structured approaches that improve solution quality systematically until target criteria is satisfied [230]. Despite the rigorous and deterministic nature of sophisticated optimization algorithms [231], the solution time for a specific problem instance largely depends on the available computational resources and underlying solution tracking algorithms. Partitioning of feasible space is a commonly adopted theme for solution algorithms, particularly for problem formulations with mixed discrete and continuous variable domains that are ubiquitous in PSE [232, 233]. Optimization problems occurring in process planning and scheduling [234] generally require shorter solution times which directly correlate with the exploration of solution space until convergence.

As computational solvers employ heuristics associated with branching and cut generation for handling and partitioning the feasible solution space [235], GenAI presents an opportunity to utilize generative deep learning to assist in developing an efficient solution process. Recently, GenAI methods like LLMs have been demonstrated to exhibit efficient solution space traversal properties, enabling them to serve as optimizers [236]. Through structured instructions, LLMs can generate new solutions for an optimization problem by learning from optimization trajectories comprising past solutions owing to their emergent properties. In-context learning by prompting in instruction-tuned LLMs like ChatGPT has also been shown to improve the optimization of a manufacturing process [237]. Furthermore, mathematical LLM models like Llemma could assist in understanding and formulating optimization problems by parsing and interpreting natural language descriptions in PSE [238]. It should be noted that the use of LLMs as optimizers can exhibit challenges in scaling for large-scale optimization problems due to the large computational resource utilization for inference with LLMs. Additionally, performance of such optimizers strongly relies on the quality and design of the engineered prompts [236] and may limit their broader applicability across optimization problems in PSE. Apart from problem formulation and iterative optimization, generative deep learning-based approaches are capable of guiding the branching heuristics in tree-based optimization algorithms for combinatorial problems [239].

GNNs form the building block of several deep learning-based approaches to guide the branching heuristics in tree-based optimization algorithms for combinatorial problems [239]. With GNNs, GenAI in combinatorial optimization has also enabled learning to branch either by imitating expert-designed branching heuristics in an RL setting [240-242] or by analyzing the likelihood of optimality with unsupervised learning [243, 244]. Owing to the ease of representation of mixed-integer programs as graphs, GNNs trained in similar settings as that of learning to branch can be exploited to generate cutting planes for bound improvements on the optimal solution value [245]. As cutting plane generation can be a challenging task for GenAI methods which do not rely on mixed-integer programs for identifying cutting planes, deep learning techniques are more suited to rank the selection of a cutting plane [246] for faster convergence of branch-and-cut based algorithms in mixed-integer programming [247, 248]. An example is G2MILP, which uses masked VAEs to augment limited data in MILP instances represented as weighted bipartite graphs [249]. With the improvements in diffusion models for data generation, graph diffusion models that operate on graph-structured data can be easily incorporated into cut generation for mathematical optimization [250]. Sequence-based architectures like long short-term memory (LSTM) have also been shown to improve cut selection for solving mixed-integer linear problems [251]. GenAI guided exploration of solution space has triggered investigations into algorithmic improvements for optimization problems like travelling salesman [252] and vehicle routing [253] which form the basis of several problems in PSE ranging from planning and scheduling [254] to supply chain optimization [255]. Combining various modalities like text in problem description and graphs for problem representations present interesting opportunities to explore the potential of GenAI models like multimodal LLMs and diffusion models for facilitating learning-based optimization in PSE. Deliberation over the benefits of incorporating GenAI into mathematical programming solution approaches and associated limitations like the computational costs of querying deep learning models within exact combinatorial optimization solvers can be required to derive practical advantages of learning to optimize for PSE problems.

*4.3. Surrogate-based optimization*

Surrogate models are approximate models used in surrogate optimization to enable efficient exploration of the solution space for optimization problems with computationally expensive objective function evaluation. These models are invaluable tools that assist in overcoming issues like computational intractability associated with analysis and optimization of complex systems in PSE [256, 257]. Although the development of tailored approaches to address large-scale PSE problems is prominent [258], surrogate-based optimization is particularly applied in case of lack of closed-form relationship between objective and variables or computationally expensive objective function evaluation [259]. ML techniques have facilitated data-driven surrogate modeling through simulated or experimental data [257, 260]. Selecting data-driven surrogate models for integration into mathematical optimization not only depends on their optimization properties but also on the sample complexity of the underlying ML algorithm [257, 261]. Deep learning-based surrogate models have been extensively utilized for optimization problems arising in PSE domains of design and control [262-264]. In addition to a tailored setting, deep learning models as surrogates have been applied to general-purpose optimization algorithms for mathematical programming formulations [265, 266].

With an increase in the adoption of deep learning in surrogate-based optimization and the need to quantify uncertainty of surrogate models [261], GenAI presents several promising avenues for integration into mathematical optimization owing to robustness towards data uncertainties and generalized predictive capabilities. GenAI models like LLMs can enable the creation of adaptive surrogate models that can learn and update their representation based on new data [267]. With diffusion-based models leveraging methods to control the information flow, exploration of the solution space with surrogate models could be potentially regulated. This approach combined with multimodal learning, can also provide for incorporating diverse input data for more accurate and robust data-driven surrogate modeling [268], especially when dealing with complex and multidimensional optimization problems arising in PSE. Time-varying trajectory modeling for complex dynamic systems is well-suited for surrogate-based optimization with sequence models like gated RNNs and LSTMs [30]. GenAI techniques using autoencoders can be leveraged by feedforward neural networks, convolutional neural networks (CNNs), as well as LSTMs to extract low dimensional representations of nonlinearities present in dynamic system responses [269, 270].

Although deep learning-based surrogate optimization exhibits potential to learn from existing physical models for predicting unknown process behaviours [271], it is important to consider the alignment of data-driven surrogates with that of the larger decision-making problems in PSE [272, 273]. To accommodate the constraints posed by the optimization problem with GenAI methods, physics-informed neural networks can incorporate model-based information to guide the learning process [274, 275]. Tailored deep learning models with nonlinear activations trained in a generative manner also exhibit potential for embedding into larger optimization problems in PSE owing to the development of mixed-integer programming formulations capable of capturing the spanned feasible space [276, 277]. LLMs and multimodal learning can further facilitate transfer learning in surrogate-based optimization [278], wherein pre-trained models can be fine-tuned on related optimization tasks to provide good quality initial candidates that directly affect the optimization trajectory. Bayesian approaches like Gaussian processes have been widely adopted as surrogate models in optimization as they express data efficiency and uncertainty quantification properties [279, 280]. Such advantages can be translated with GenAI for surrogate-based optimization in PSE with the integration of Bayesian approaches into deep learning with strategies like deep Gaussian processes and Bayesian learning [281, 282]. Bayesian optimization supplemented by generative LLMs can also enable iterative promising solution suggestions conditioned on historical evaluations through in-context learning. Such GenAI technologies offer opportunities to leverage meta-surrogate optimization to enhance the adaptability, efficiency, and accuracy of surrogate models, making them valuable tools in addressing complex optimization problems in PSE. It is important to note that the robust nature of the underlying surrogate model in response to changes in problem data or input conditions must be ensured by constraining the surrogate models to avoid large extrapolation errors [261]. Another challenge lies in verifying that optimal solutions of GenAI-enhanced surrogate models correspond to those of original optimization problem and is an important factor that must be addressed before adopting GenAI into surrogate-based optimization for large-scale PSE problems.

### 4.4. Metaheuristic optimization

Metaheuristic algorithms provide a trade-off between local search and randomization for optimization problems in PSE spanning across domains of design and synthesis, scheduling and planning, as well as control and monitoring [283]. In addition to tackling domain-specific problems, as metaheuristic algorithms generally tend to be suitable for global optimization [284], they possess the potential to tackle optimization under uncertainty [285, 286], optimizing surrogates in surrogate-based optimization [287], as well as multi-objective optimization [288, 289] for various PSE applications. As evaluating the fitness function which dictates the selection criterion for a particular problem is critical for efficient exploration with the employed metaheuristic, there is a prevailing need for integration with data-driven techniques in case of computationally expensive fitness evaluations for many real-world optimization problems [290]. A natural extension of GenAI into metaheuristic optimization is the use of deep learning-based models as surrogates for data-driven fitness function estimation and sampling [291, 292]. On the other hand, metaheuristic algorithms can also be exploited for generative deep learning models for various PSE applications to overcome limitations like computationally expensive convergence to local minima associated with conventional training methods [293, 294].

The forecasting capabilities of GenAI methods leveraging sequence models like LSTMs and Transformer can be integrated with a metaheuristic engine for time-series data [295]. Solving nonlinear optimal control and multi-period planning problems that require future system behavior predictions can benefit from such hybrid methods and have been explored with evolutionary metaheuristic algorithms [296, 297]. Solution techniques for multi-objective optimization for PSE problems like reactor design and process scheduling have seen the use of deep learning models for fitness function approximations [298, 299], and demonstrate potential for adopting GenAI methods to incorporate domain-specific knowledge into the solution exploration strategy [300]. Conversely, metaheuristics like evolutionary and swarm intelligence techniques can guide the design of generative deep learning models and corresponding hyperparameters through sampling and fitness evaluation of feasible architecture designs and their performance [301]. As generative training of deep learning models is inherently a large-scale optimization problem, metaheuristic optimization promotes a gradient-free approach that can aid in obtaining robust set of parameters in a computationally inexpensive manner [302, 303]. A promising direction where metaheuristic optimization can be supplanted by advanced GenAI methods like LLMs is agent-based modeling in PSE applications [304]. Problems like supply chain optimization can be tackled with agent-based model comprising a collection of decision-making entities followed by solution with metaheuristics like genetic algorithms [305]. LLMs incorporating multiple modalities can serve as agents in such frameworks and can generalize to various tasks across multiple domains in PSE [306]. Even though the adoption of deep learning with metaheuristics for practical industrial problems remains sparse [293], the interplay between GenAI and metaheuristic optimization presents an emerging opportunity for developing data-driven computationally efficient strategies to address PSE problems at various scales.

## 5. Generative AI for Process Monitoring and Control

Since its inception, process monitoring and control (PMC) has been fundamental in PSE, primarily concentrating on ensuring efficiency, safety, and compliance with environmental regulations [3, 307]. Modern process control systems not only continuously track and adjust key process variables such as temperature, pressure, and chemical composition, to boost yield, enhance efficiency, and improve product quality, but also now consider economic information in real time [308, 309]. Process monitoring systems also play a critical role in maintaining quality and safety by detecting and mitigating hazardous conditions early, thereby aligning with stringent safety standards and averting potential accidents [310]. Over time, PMC has evolved to include optimal strategies, addressing more complex and large-scale challenges propelled by the increasing need for sustainable production processes that are both cleaner and more efficient, reducing resource use and waste. Addressing the increased complexity of these systems requires innovative solutions different from the traditional approaches, for which reason GenAI emerges as an important technology for achieving these advanced objectives.

In this section, we explore the potential of GenAI to advance PMC to achieve the increased objectives of PSE, drawing inspiration from neighboring domains where GenAI has had tremendous success. We restrict the discussion to typical activities integral to PMC including modeling, control, sensing, and fault detection as shown in Figure 4. These activities, although seemingly disparate, are strongly interconnected. Process modeling provides a mathematical framework for understanding and simulating processes, forming the basis for effective control strategies. Process control uses these models to regulate and maintain operational variables. Soft sensing, an extension of process control, uses models to estimate unmeasurable or costly-to-measure process variables. Fault detection relies on both process models and control feedback to identify anomalies. Thus, each component, while distinct, feeds into and enhances the others, creating a cohesive system for optimal process management. We begin the section with dynamic process modeling, and then proceed to soft sensing. Thereafter, we move on to process control and finally conclude the section with fault detection and identification. It is worth noting that the analysis here is not to provide an extensive review of the techniques in these areas in PMC, but to examine the potential of GenAI to impact the activities in PMC.

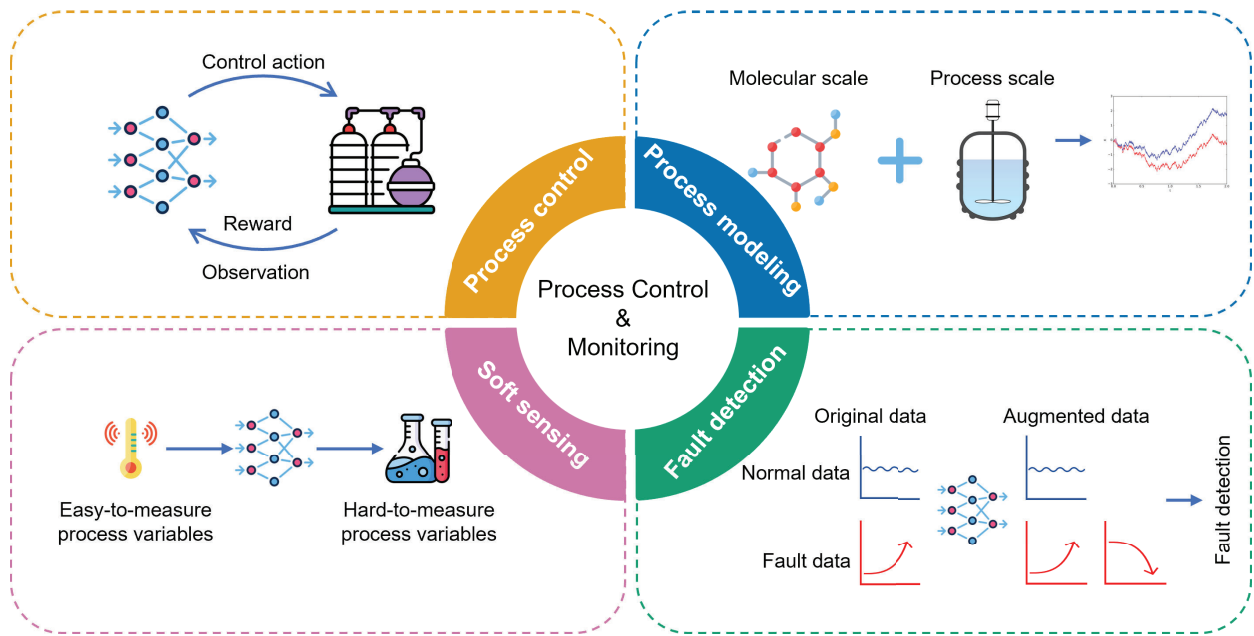

Figure 5. GenAI enhances process monitoring and control in several ways: it improves process control using RL, models complex dynamics across scales, enhances soft sensing for quick estimation of hard-to-measure variables, and augments data for better fault detection and identification.

### 5.1. Modeling process dynamics for control

Dynamic process models play an important role in control systems, primarily because they allow for the prediction of system behavior under a variety of conditions over time. These dynamic models are generally developed based on first principles or through data-driven approaches, tailored for control purposes. Among them, linear dynamic models have been extensively studied for their simplicity and computational efficiency [311], although they struggle to accurately capture the dynamic and nonlinear characteristics of complex process systems [14]. This difficulty is compounded when attempting to create a first-principles dynamic model for large-scale processes with complex, nonlinear behaviors, leading to a historical emphasis on data-driven modeling. To address these challenges, several methods have been developed. For instance, nonlinear subspace system identification methods have been used to identify state-space dynamic models [312, 313]. Similarly, the nonlinear autoregressive moving average with exogenous input (NARMAX) method has been introduced for identifying nonlinear dynamic input-output models [314]. These data-driven dynamic models significantly benefit from efficient data planning and collection for system identification. Advanced tools like GenAI, including multimodal LLMs such as GPT-4 [20], can be instrumental in conducting exploratory analysis on extensive process data [315]. This analysis is critical to unravel complex nonlinear relationships among multiple process variables and to select appropriate input-output pairs for system identification and process control. LLMs particularly ChatGPT have already proven effective in autonomously designing, planning, and conducting experiments in chemical research [316]. LLMs therefore hold potential for planning experiments to gather rich and diverse data necessary for system identification, ultimately reducing the time required to identify dynamic process models for control and estimation. Yet, the application of LLMs in analyzing process data to identify input-output variables and in planning experiments for system identification remains an underexplored area.

Moving away from traditional identification methods, GenAI holds the potential to impact the modeling of process dynamics in process control, offering innovative solutions to overcome the limitations of traditional identification methods. Owing to their ability to model time-series data, RNNs, and its variants, have proven to be successful in approximating nonlinear dynamical systems [68, 317, 318] and systems that exhibit both continuous and discrete dynamics [319]. However, these dynamic models are deterministic and limited within the data used to develop them. By integrating stochastic elements at the output level and substituting the deterministic output data sequence with a sequence of distribution parameters, RNNs can be effectively converted into generative RNNs (GRNNs) through the definition of a probabilistic observation model [320]. Going forward, the use of GRNNs can be used to understand and mimic the underlying probability distributions of the training data helping to increase the dynamic model performance for extrapolation in unseen or even dangerous system behavior. Another successful GenAI model for modeling time-series data with demonstrated superiority in terms of fewer parameters, and 15 % reduced computation time over standard RNNs in a batch crystallization process is Transformers [321, 322]. This success is attributed to the effectiveness of the attention mechanisms inherent in Transformers. Considering that ML-based dynamic process models may require a high amount of data for training, the synthetic data generation and augmentation capabilities of GANs can be harnessed to obtain hard-to-collect data or reduce the data requirements. Furthermore, multimodal LLMs, GANs, and the few-shot learning framework [323] may be combined into a holistic framework to plan minimum data collection experiments for model training. The potential of GenAI to further improve modeling process dynamics for control and estimation leaves much room to be explored.

*5.2. Soft sensing and state estimation*

Soft sensing and state estimation are important in process control for providing real-time, cost-effective monitoring of hard-to-measure variables, thereby enhancing safety, quality, and efficiency in complex industrial processes. These techniques rely on predictive models that estimate the unmeasured states of a process based on available sensor data, thereby facilitating real-time monitoring and control [324]. However, implementing these methods effectively comes with its challenges. One significant issue is the potential inaccuracy of dynamic models used for estimation [325]. These dynamic models often rely on assumptions that may not hold true under varying process conditions, leading to errors in estimation. Another challenge is the presence of noise and uncertainty in sensor data, which can degrade the quality of the state estimates [326]. Furthermore, processes often evolve over time due to changes in raw materials, environmental conditions, or equipment wear, necessitating adaptive dynamic models that can adjust to these changes [327, 328].

GenAI, with its ability to learn complex patterns and dynamics from data, can significantly aid in soft sensing and state estimation. VAEs can take into account the presence of uncertainties in process variables for better just-in-time adaptive soft sensing [326], and fill in gaps in data [329], leading to more reliable state estimates. Although the success of VAE in deep feature extraction and uncertain data modeling is notable, its instability and reconstruction errors, arising from the random sampling process in the latent subspace that represents the original input space is an issue. This challenge can be mitigated using the Constrained VAE soft sensing approach [330]. RL has been used to address cross-domain soft sensor problem to overcome the lack of reliability of soft sensors in new operating conditions [331]. The RL-based soft sensing approach achieved an RMSE of 0.1201 compared to existing approaches in an industrial steam-assisted gravity drainage (SAGD) process. The growing reliance on images for capturing spatial information for process control necessitates advanced segmentation capabilities and extensive training data for ML models. GANs and their variants, renowned for producing realistic synthetic images, are key in generating additional data, particularly in scenarios with limited or hard-to-collect data. For example, most data primarily in PSE represents normal operating conditions, which leads to a data imbalance known as the "small data" problem. These synthetic images enhance datasets, tackling the "small data" challenge by increasing the volume and diversity of training samples [332-334]. The segment anything model (SAM) [335] could further strengthen this image approach with its ability to precisely segment a wide array of objects within images, irrespective of specific training thus enhancing image-based soft sensing. Combining GANs with SAM can create a powerful synergy. While GANs enhance the dataset by adding synthetic yet realistic images, SAMs ensure accurate segmentation across a broad spectrum of objects and scenarios. Furthermore, the integration of GenAI models like CLIP could enhance image-based sensing with its capabilities in zero-shot transfer, multimodal learning, and natural language supervision. These advancements promise significant improvements in fields like automated control systems and environmental monitoring, showcasing GenAI's potential in improving sensing and estimation methods.

### 5.3. Process control

Modern process controllers are tasked with maintaining operational safety while simultaneously handling multiple objectives, such as economic efficiency and sustainability, at various scales [336]. This multifaceted objective significantly increases the complexity of PSE techniques used in process control. One significant challenge in process control is dealing with complex, multivariable systems where interactions between different process variables can be complex and unpredictable [337]. Traditional control strategies might struggle to cope with such complexity, leading to suboptimal performance or even safety risks. Another challenge arises from the need to adapt to changing external conditions and raw material qualities, requiring a flexible and responsive control system [338]. Considering that process control is central to achieving PSE objectives, it is important to explore GenAI to improve the performance of process control.

GenAI has emerged as a tool for addressing the complexities of process control. Transformers, known for their enhanced accuracy and efficiency, can serve as surrogate models in model-based control, thereby enhancing robustness and reducing computational time [16, 321, 339]. Furthermore, VAEs has been used to analyze vast amounts of process data and trajectories to identify safe sets in robotic systems for food-cutting [340]. As processes grow in scale and complexity, integrating diverse economic and sustainability objectives, RL emerges as a particularly promising strategy for process control [341]. RL can select optimal actions under uncertain conditions significantly boosting both responsiveness and robustness of process control, while also curtailing online computation time. RL has also shown promise in automatically fine-tuning proportional-integral-derivative (PID) controllers, thereby enhancing their adaptability and performance [342-344]. Furthermore, DeepMind's MuZero [99] and Gato [23] stand out with their superior multimodal and few-shot learning abilities, making them valuable tools in the use of multimodal process control such as image-based control. While these tools are currently underexplored in PSE, this approach could significantly expedite controller design, bypassing the lengthy process of developing dynamic process models for current advanced process control methods. However, there is also a growing need to focus more on safety aspects, like ensuring RL decisions consistently satisfy operational constraints. Finally, multimodal LLMs like GPT-4 offer a unique perspective. GPT-4 can be used to analyze process and instrumentation diagrams (P&IDs) to detect potential flaws in the control systems design and make necessary recommendations. This capability highlights the diverse applications of GenAI in enhancing the reliability and efficiency of process control systems.

### 5.4. Fault detection and identification

Rapid detection and accurate identification of faults in control systems are critical for ensuring their safety and reliability, which helps prevent equipment damage, minimize downtime, maintain product quality, and protect human lives [345]. One of the key challenges in fault detection and identification (FDI) is the complexity of modern process systems, which often contain a vast array of interconnected components and variables. Detecting faults in such a complex system can be akin to finding a needle in a haystack, especially when the symptoms of a fault are subtle or masked by other operational variations. Additionally, the identification of the specific type and location of a fault is complicated by the interdependencies within the system, where a single malfunction can have cascading effects.

GenAI, with its ability to learn complex patterns from data, has the capability to improve FDI methods. Common approaches encompass using VAEs [44, 45] to discern complex patterns and relationships in process data, or feature extraction–sometimes through generative training enhanced with quantum sampling [346] or dimensionality reduction with statistical approaches [347] to enable visual representations of the problem space. These approaches, which aid in the identification of potential anomalous clusters, typically entail modeling historical anomalies, feature identification, and verifying internal consistencies or inconsistencies. However, these techniques require frequent retraining and fine-tuning of numerous parameters, making the process slow. LLMs such as GPT-4 could be used to address these issues in fault detection. This use of LLMs presents significant advantages as its implementation is straightforward while eliminating the need for further model training due to their zero-shot learning ability. Furthermore, utilizing a pre-trained LLM circumvents the need for extensive parameterization, feature engineering, and specific model training for fault detection. The use of LLMs in PSE for fault detection and identification has however not been explored. In another direction, GANs can be used to generate synthetic data of faulty scenarios, which can augment the existing process data to address the "small data" problem, thereby enhancing the robustness and effectiveness of fault detection and identification models [29, 348]. An additional promising area is the use of RL to aid in fault identification by assigning agents to predict fault accurately while learning from the dynamic process environment and showcasing 93 % mean accuracy in machine tools [349].

# 6. Generative AI Challenges in PSE

While the integration of GenAI into PSE is a promising venture, it is not without challenges. An overview of the challenges is presented in Figure 5. In this section, we discuss potential challenges that may be encountered and must be addressed for PSE to fully adopt GenAI into its solution methods.

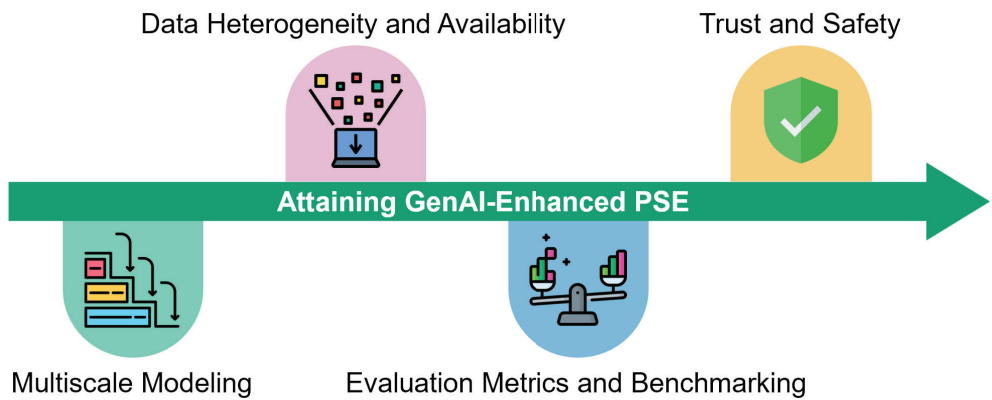

Figure 6. Potential challenges that needed to be addressed for effective and full adoption of GenAI models into the core PSE domains: synthesis and design, optimization and integration, and process monitoring and control. The key challenges include addressing modeling scale modeling, data heterogeneity and availability, effective evaluation benchmarking and metrics and ensuring trust and safety in the GenAI solution.

### 6.1. Addressing Multiscale Modelling

Multiscale modeling in PSE presents a complex hurdle for GenAI application as it requires the integration of data across various scales from molecular to enterprise. A major issue that needs to be addressed under multiscale modeling, has to do with developing consistent framework for coherent and easy representation of data across the different scales [350]. For instance, LLMs, including GPT-4 which powers ChatGPT, are generalists by nature and may lack the deep, specific domain expertise required for accurate integration or analysis within the context of multiscale modeling. Therefore, more specialized versions of GenAI models for multiscale modeling are desirable. However, this will require a significant amount of data, training times and associated training costs. A possible solution to mitigate this issue is to finetune LLMs specifically to handle multiscale modeling problems often encountered in PSE. In this regard, transfer learning will be a useful approach to pursue [351].

Furthermore, using RNNs or their variants and Transformers in dynamic multiscale modeling within PSE presents distinct challenges. A primary challenge is the inherent complexity of these models in capturing the dynamics across different scales, from molecular to macroscopic processes. This complexity can lead to difficulties in training the models effectively, especially when dealing with long sequence dependencies typical in process systems. A possible solution is to develop hybrid dynamic models that combine these autoregressive models with physics information to help capture and understand the long sequence dependencies. These challenges highlight the need to tailor efforts to address the challenges on integrating GenAI into multiscale PSE problems.

### *6.2. Data Heterogeneity and Availability*

GenAI in PSE, like GenAI in other disciplines grapples with challenges in data representation [352, 353], requiring significant efforts. PSE problems span broad disciplines and application areas which may require tailored solutions specific to those problems, leading to heterogenous data that needs significant processing, particularly for multimodal GenAI types such as GPT-4, CLIP, and Gato. The heterogeneity and availability of PSE data call for improved data integration and collaborative sharing to facilitate efficient integration of these models in current PSE methods. Implementing extensive data normalization techniques can make the data more homogenous and digestible [354]. Furthermore, the availability of quality, relevant data is a critical challenge. PSE data may be scarce [355], fragmented [356], or inconsistent [357], which hampers the training and performance of GenAI models.

Another problem is data imbalance or distribution [358] which is prevalent in PSE data. For instance, most operating data collected in manufacturing plants falls within the normal operating regime, with less data on faulty abnormal scenarios. Using this data to train GenAI models like CLIP, GPT, Gato, despite their multimodality and few-shot or zero-shot learning capabilities may exhibit bias in their generation capabilities and ultimately underperform. To address this challenge, generative models can be used to augment the existing datasets thus reducing the need to not only collect extensive amounts of data, but also label them as ground truth. Another solution is to harmonize data into a comprehensive data sharing framework led by PSE researchers and practitioners. These frameworks allow for standardized data sets which are valuable in evaluating the performance of different GenAI models. It is important to note that industrial PSE data is often confidential, which can significantly hinder collaborative data sharing efforts.

*6.3. Evaluation Metrics and Benchmarks*

Evaluating and benchmarking GenAI models, such as LLMs, image generation models, and multimodal models, presents a set of complex challenges, especially in the field of PSE. Each model type serves different purposes and thus necessitates specific evaluation metrics to gauge their effectiveness and reliability. For instance, traditional NLP metrics like bilingual evaluation understudy (BLEU) [359] or F1 scores are insufficient in PSE contexts due to their focus on general language accuracy, failing to capture the specialized vocabulary, contextual relevance, and high precision required in PSE. In the field of PSE, it is critical to assess both linguistic precision and the model's relevance to the specific tasks. In terms of image generation models, their evaluation in PSE should emphasize not just the visual appeal but also the technical accuracy and practicality of the generated images. While metrics such as Inception Score (IS) [360] or Frechet Inception Distance (FID) [361] are useful in PSE to determine the quality of images generated, these metrics must be supplemented with domain-specific evaluations to check the utility of the images. The Kernel Inception Distance (KID) [362] metric offers a more comprehensive evaluation by measuring differences in the feature space of generated and real images. However, it has limitations, like difficulty in detecting overfitting and inapplicability to non-visual data, encountered in some VAE applications.

For multi-modal models like Gato, the evaluation in PSE should cover a broad spectrum of metrics reflecting its diverse capabilities. For example, in process control, the model's precision in manipulating control variables, adaptability to changing conditions, and efficiency in learning new tasks are important. The primary challenge in evaluating these models is finding metrics that not only measure technical performance (such as accuracy, precision, or speed) but also consider domain-specific factors like safety, reliability, and compliance with industry standards. Furthermore, these models should be evaluated for their ability to generalize across various tasks while maintaining high performance, which is critical in the varied and unpredictable PSE domain. Integrating various metrics like IS, FID, and KID could offer a more comprehensive evaluation perspective. To conclude, assessing GenAI in PSE requires a multifaceted approach that includes traditional performance metrics, domain-specific assessments, interpretability, and real-world applicability. This remedy guarantees that these advanced technologies meet the unique demands of PSE.

*6.4. Trust and Safety*

Ensuring the reliability and safety of GenAI-generated solutions, particularly in safety-critical applications, is a significant and multifaceted challenge due to the complexity of process systems, data quality issues, the need for high model accuracy, and stringent regulatory standards. Similar challenges occur in domains such as healthcare [363]. The complexity inherent in PSE systems requires GenAI models such as LLMs and their multimodal counterparts to be deeply integrated with chemical, physical, and PSE principles to produce safer and more reliable outcomes. However, this integration is complex due to the interactions and processes involved in PSE. LLMs have been found to hallucinate quite often, producing information which are false [364]. This false information could be fatal in critical PSE applications and necessitates the integration of human expertise in the evaluation loop, together with rigorous evaluation and benchmarking metrics as discussed in the preceding subsection. For instance, if a GenAI model in a chemical plant inaccurately predicts unsafe reaction conditions, such as excessively high temperatures, it could cause a catastrophic event, leading to material damage, financial loss, and severe risks to human safety and the environment.

Additionally, a major challenge lies in the "black box" nature of many AI models including GenAI ones [365]. The lack of transparency in how these models arrive at decisions hinders trust among users, which is important in a field as critical as PSE. For example, the use of Gato to make complex decisions in real-time process control would require strong guarantees that the decisions are within the safe operating parameters. A potential way to address this challenge is to integrate PSE domain knowledge directly into these cutting-edge GenAI models during training to enhance their reliability. Integrating PSE domain knowledge into GenAI models like LLMs and Gato can be achieved through various methods, each suited to different scenarios. Incorporating this knowledge during the training process allows the model to learn directly from domain-enriched data, while transfer learning is effective for adapting a pre-trained model to PSE specifics, offering efficiency in cases of limited domain data. In-context learning, on the other hand, is flexible and useful for dynamic adjustment of the model's responses, particularly beneficial for models capable of understanding and responding to specific contextual cues in PSE tasks. This approach might involve creating hybrid models that blend traditional PSE methods with GenAI models, ensuring that the outputs are grounded in established scientific principles. Another key solution is the advancement of explainable AI (XAI) [366]. Developing and integrating XAI techniques can make AI models more transparent and understandable. Techniques like feature importance scoring [367], decision trees [368], or visualizations could provide valuable insights into the AI decision-making process. Robust testing and validation protocols are also critical to guarantee the reliability of GenAI models. Finally, collaboration with regulatory bodies is important to check that practical GenAI solutions meet all safety, cybersecurity, and compliance requirements.

### 6.5. Future expectations

The complexity and variability of industrial processes call for sophisticated multiscale modeling techniques that can capture the complex nonlinear behavior in chemical and biological processes, from molecular phenomenon to enterprise-wide supply chain analytics. In this regard, future developments in GenAI methods, particularly foundation models and LLMs, trained on rich PSE domain knowledge to capture this multiscale complexity is desired. Additionally, a seamless integration of well-established PSE tools based on first principles and GenAI must be explored. This integration will not only enhance the robustness of GenAI models, but also enhance their explainability which is a key metric in PSE. However, interdisciplinary collaboration is important for these developments to mature from theoretical models to practical applications. Collaborations between AI experts, and chemical engineering practitioners and researchers are essential to tailor GenAI tools that not only enhance design methods, but also operational efficiency while adhering to high safety standards. Furthermore, the development of open platforms where researchers and industry experts from diverse domains in PSE can share data, tools, and insights will also play a critical role. This approach fosters a community-driven development environment where challenges such as data heterogeneity and model transparency can be tackled more effectively. Creating these collaborative ecosystems will not only speed up the adoption of GenAI solutions in PSE but also ensure that these technologies are developed in an ethical and socially responsible manner.

## 7. Concluding Remarks

Recent advancements in GenAI, including LLMs like GPT, image generation models like Stable Diffusion, and multimodal decision-making models like Gato have opened new frontiers in many fields of science and engineering. The advancements in GenAI models therefore hold great promise to advance the solution methods of PSE. In this review paper, we explored the significant advancements in GenAI, as well as their opportunities and challenges within the context of PSE. We began by offering a comprehensive overview of GenAI models. This encompassed both classic models (such as autoencoders, autoregressive, and generative adversarial network models) and state-of-the-art technologies (like GPT, Stable Diffusion, CLIP, and Gato). Our analysis highlighted their unique features and their relevance in PSE applications. Subsequently, we provided a detailed survey of how GenAI models are being applied within key PSE domains: synthesis and design, optimization and integration, and process monitoring and control. For each domain, we also discussed the potential of these advanced GenAI technologies. We particularly emphasized the role of multimodal LLMs in enhancing solution strategies in PSE. The paper also addressed the challenges that need to be tackled for the effective integration of GenAI in PSE. These challenges included aligning GenAI capabilities with the multiscale needs of PSE, ensuring robustness for greater safety and trust, managing data availability and heterogeneity, and developing relevant evaluation benchmarks and metrics. As the field of PSE continues to evolve, embracing the innovative solutions offered by GenAI will be critical in tackling complex global challenges. This paper serves as a foundation for future research and development, encouraging continued exploration and refinement of GenAI applications in PSE, thereby paving the way for more sustainable, efficient, and intelligent process systems.

## Acknowledgements

B.D.-N. acknowledges the partial support from Schmidt Futures via an Eric and Wendy Schmidt AI in Science Postdoctoral Fellowship to Cornell University. We also acknowledge the use of generative AI to polish the language in the manuscript. Figures are created with BioRender.com.

[221] A. Alcántara and C. Ruiz, "On data-driven chance constraint learning for mixed-integer optimization problems," *Applied Mathematical Modelling,* vol. 121, pp. 445-462, 2023.

[222] M. Goerigk and J. Kurtz, "Data-driven robust optimization using unsupervised deep learning," *arXiv preprint arXiv:2011.09769,* 2020.

[223] A. R. Chenreddy, N. Bandi, and E. Delage, "Data-driven conditional robust optimization," *Advances in Neural Information Processing Systems,* vol. 35, pp. 9525-9537, 2022.

[224] C. Wang, X. Peng, C. Shang, C. Fan, L. Zhao, and W. Zhong, "A deep learning-based robust optimization approach for refinery planning under uncertainty," *Computers & Chemical Engineering,* vol. 155, p. 107495, 2021/12/01/ 2021, doi: https://doi.org/10.1016/j.compchemeng.2021.107495.

[225] M. Goerigk and J. Kurtz, "Data-driven robust optimization using deep neural networks," *Computers & Operations Research,* vol. 151, p. 106087, 2023/03/01/ 2023, doi: https://doi.org/10.1016/j.cor.2022.106087.

[226] B. Song, R. Zhou, and F. Ahmed, "Multi-modal machine learning in engineering design: A review and future directions," *Journal of Computing and Information Science in Engineering,* vol. 24, no. 1, p. 010801, 2024.

[227] S. Bubeck *et al.*, "Sparks of artificial general intelligence: Early experiments with gpt-4," *arXiv preprint arXiv:2303.12712,* 2023.

[228] Z. Niu, G. Zhong, and H. Yu, "A review on the attention mechanism of deep learning," *Neurocomputing,* vol. 452, pp. 48-62, 2021.

[229] I. E. Grossmann, J. A. Caballero, and H. Yeomans, "Advances in mathematical programming for the synthesis of process systems," *Latin American Applied Research,* vol. 30, no. 4, pp. 263-284, 2000.

[230] D. P. Bertsekas, "Nonlinear programming," *Journal of the Operational Research Society,* vol. 48, no. 3, pp. 334-334, 1997.

[231] R. Horst and H. Tuy, *Global optimization: Deterministic approaches*. Springer Science & Business Media, 2013.

[232] I. E. Grossmann and Z. Kravanja, "Mixed-integer nonlinear programming techniques for process systems engineering," *Computers & chemical engineering,* vol. 19, pp. 189-204, 1995.

[233] J. Kallrath, "Mixed integer optimization in the chemical process industry: Experience, potential and future perspectives," *Chemical Engineering Research and Design,* vol. 78, no. 6, pp. 809-822, 2000.

[234] I. E. Grossmann, S. A. Van Den Heever, and I. Harjunkoski, "Discrete optimization methods and their role in the integration of planning and scheduling," in *AIChE Symposium Series*, 2002: New York; American Institute of Chemical Engineers; 1998, pp. 150-168.

[235] T. Berthold, *Heuristic algorithms in global MINLP solvers*. Verlag Dr. Hut, 2015.

[236] C. Yang *et al.*, "Large language models as optimizers," *arXiv preprint arXiv:2309.03409,* 2023.

[237] S. Badini, S. Regondi, E. Frontoni, and R. Pugliese, "Assessing the capabilities of ChatGPT to improve additive manufacturing troubleshooting," *Advanced Industrial and Engineering Polymer Research,* vol. 6, no. 3, pp. 278-287, 2023/07/01/ 2023, doi: https://doi.org/10.1016/j.aiepr.2023.03.003.

[238] Z. Azerbayev *et al.*, "Llemma: An open language model for mathematics," *arXiv preprint arXiv:2310.10631,* 2023.
64